\documentclass{article}
\usepackage{ijcai19}

\usepackage{times}
\usepackage{soul}
\usepackage{url}
\usepackage[hidelinks]{hyperref}
\usepackage[utf8]{inputenc}
\usepackage[small]{caption}
\usepackage{graphicx}
\usepackage{amsmath}
\usepackage{booktabs}
\usepackage{algorithm}
\usepackage{algorithmic}
\title{FedHealth: A Federated Transfer Learning Framework for Wearable Healthcare}

\author{
Yiqiang Chen$^{1,2,3}$\footnote{Corresponding Author}\and
Jindong Wang$^4$\and
Chaohui Yu$^{1,2}$\and
Wen Gao$^{3}$\and
Xin Qin$^{1,2}$\\
\affiliations
$^1$Beijing Key Lab. of Mobile Computing and Pervasive Devices, Inst. of Computing Tech., CAS\\
$^2$University of Chinese Academy of Sciences, Beijing, China\\
$^3$Pengcheng Laboratory, Shenzhen, China\\
$^4$Microsoft Research Asia, Beijing, China
\emails
yqchen@ict.ac.cn, jindong.wang@microsoft.com
}

\begin{document}

\maketitle

\begin{abstract}
With the rapid development of computing technology, wearable devices such as smart phones and wristbands make it easy to get access to people's health information including activities, sleep, sports, etc. Smart healthcare achieves great success by training machine learning models on large quantity of user data. However, there are two critical challenges. Firstly, user data often exists in the form of isolated islands, making it difficult to perform aggregation without compromising privacy security. Secondly, the models trained on the cloud fail on personalization. In this paper, we propose FedHealth, the first federated transfer learning framework for wearable healthcare to tackle these challenges. FedHealth performs data aggregation through federated learning, and then builds personalized models by transfer learning. It is able to achieve accurate and personalized healthcare without compromising privacy and security. Experiments demonstrate that FedHealth produces higher accuracy (5.3\% improvement) for wearable activity recognition when compared to traditional methods. FedHealth is general and extensible and has the potential to be used in many healthcare applications.
\end{abstract}

\section{Introduction}

Activities of daily living (ADL) are highly related to people's health. Recently, the development of wearable technologies helps people to understand their health status by tracking activities using wearable devices such as smartphone, wristband, and smart glasses. Wearable healthcare has the potential to provide early warnings to several cognitive diseases such as Parkinson's~\cite{chen2017pdassist,chen2019cross} and small vessel diseases~\cite{chen2018inferring}. Other applications include mental health assessment~\cite{wang2014studentlife}, fall detection~\cite{wang2017wifall}, and sports monitoring~\cite{wang2019deep}. In fact, there is a growing trend for wearable healthcare over the years~\cite{andreu2015wearable,hiremath2014wearable}. 

In healthcare applications, machine learning models are often trained on sufficient user data to track health status. Traditional machine learning approaches such as Support Vector Machines~(SVM), Decision Tree~(DT), and Hidden Markov Models~(HMM) are adopted in many healthcare applications~\cite{ward2016towards}. The recent success of deep learning achieves satisfactory performances by training on larger sizes of user data. Representative networks include Convolutional Neural Networks~(CNN), Recurrent Neural Networks~(RNN), and Autoencoders~\cite{wang2019deep}. 

\begin{figure}[t!]
	\centering
	\includegraphics[scale=.54]{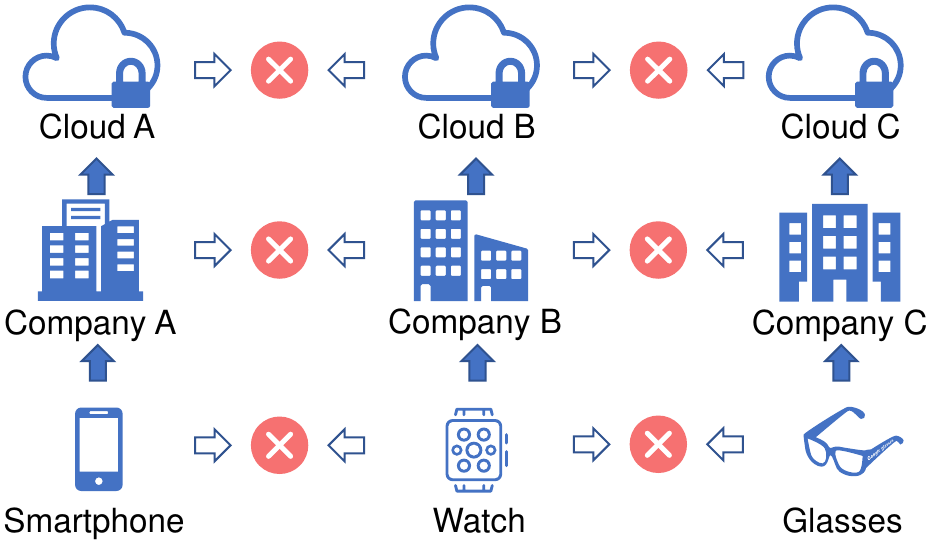}
	\caption{The data islanding and personalization problems in wearable healthcare}
	\label{fig-intro}
\end{figure}

Unfortunately, there are two critical challenges in today's wearable healthcare (Figure~\ref{fig-intro}). First of all, in real life, data often exists in the form of isolated islands. Although there are plenty of data in different organizations, institutes, and subjects, it is not possible to share them due to privacy and security concerns. In Figure~\ref{fig-intro}, when the same user uses different products from two companies, his data stored in two clouds cannot be exchanged. This makes it hard to train powerful models using these valuable data. Additionally, recently, China, the United States, and the European Union enforced the protection of user data via different regularizations~\cite{inkster2018china,voigt2017eu}. Hence, the acquisition of massive user data is not possible in real applications.

The other important issue is personalization. Most of the methods are based on a common server model for nearly all users. After acquiring sufficient user data to train a satisfactory machine learning model, the model itself is then distributed to all the user devices on which the daily health information can be tracked. This process lacks personalization. As can be seen, different users have different physical characteristics and daily activity patterns. Therefore, the common model fails to perform personalized healthcare.

In this paper, we propose \textit{FedHealth}, the first federated transfer learning framework for wearable healthcare. FedHealth can solve both of the data islanding and personalization problems. Through federated learning~\cite{yang2019federated,yang2018federated} and homomorphic encryption~\cite{rivest1978data}, FedHealth aggregates the data from separate organizations to build powerful machine learning models with the users' privacy well preserved. After the cloud model is built, FedHealth utilizes transfer learning~\cite{pan2010survey} methods to achieve personalized model learning for each organization. The framework can incrementally update. FedHealth is extensible and can be deployed to many healthcare applications to continuously enhance their learning abilities in real life.

In summary, this paper makes the following contributions:

1. We propose FedHealth, the first federated transfer learning framework for wearable healthcare, which aggregates the data from different organizations without compromising privacy security, and achieves personalized model learning through knowledge transfer.

2. We show the excellent performance achieved by FedHealth in smartphone based human activity recognition. Experiments show that FedHealth dramatically improves the recognition accuracy by 5.3\% compared to traditional learning approaches.

3. FedHealth is extensible and can be the standard framework to many healthcare applications. With the users' privacy well preserved and good performance achieved, it can be easily deployed to other healthcare applications.

\section{Related Work}

In this section, we introduce the related work in three aspects: wearable healthcare, federated machine learning, and transfer learning.

\subsection{Wearable Healthcare}

Certain activities in daily life reflect early signals of some cognitive diseases~\cite{atkinson2007cognitive,michalak2009embodiment}. For instance, the change of gait may result in small vessel disease or stroke. A lot of researchers pay attention to monitor users' activities using body-worn sensors~\cite{voigt2017eu}, through which daily activities and sports activities can be recognized. With the development of wearable technology, smartphone, wristbands, and smart glasses provide easy access to this information. Many endeavors have been made~\cite{wang2014studentlife,albinali2010using,wang2017wifall}. Other than activities, physiological signals can also help to detect certain diseases. EEG (electroencephalography) is used to detect seizures~\cite{menshawy2015automatic,hiremath2014wearable}. Authors can also use RGB-D cameras to detect users' activities~\cite{lei2012fine,rashidi2010mining}. For a complete survey on sensors based activity recognition and healthcare, interested readers are recommended to refer to \cite{wang2019deep}.

It is noteworthy that traditional healthcare applications often build the model by aggregating all the user data. However, in real applications, data are often separate and cannot be easily shared due to privacy issues~\cite{inkster2018china,voigt2017eu}. Moreover, the models built by applications lack the ability of personalization.

\subsection{Federated Machine Learning}

A comprehensive survey on federated learning is in \cite{yang2019federated}. Federated machine learning was firstly proposed by Google~\cite{konevcny2016federated,konevcny2016federated}, where they trained machine learning models based on distributed mobile phones all over the world. The key idea is to protect user data during the process. Since then, other researchers started to focus on privacy-preserving machine learning~\cite{bonawitz2017practical,shokri2015privacy,geyer2017differentially}, federated multi-task learning~\cite{smith2017federated}, as well as personalized federated learning~\cite{chen2018federated}. Federated learning has the ability to resolve the data islanding problems by privacy-preserving model training in the network. 

According to \cite{yang2019federated}, federated learning can mainly be classified into three types: 1) horizontal federated learning, where organizations share partial features; 2) vertical federated learning, where organizations share partial samples; and 3) federated transfer learning, where neither samples or features have much in common. FedHealth belongs to federated transfer learning category. It is the first of its kind tailored for wearable healthcare applications.

\subsection{Transfer Learning}

Transfer learning aims at transferring knowledge from existing domains to a new domain. In the setting of transfer learning, the domains are often different but related, which makes knowledge transfer possible. The key idea is to reduce the distribution divergence between different domains. To this end, there are mainly two kinds of approaches: 1) instance reweighting~\cite{huang2012boosting,huang2007correcting}, which reuses samples from the source domain according to some weighting technique; and 2) feature matching, which either performs subspace learning by exploiting the subspace geometrical structure~\cite{wang2018visual,sun2016return,fernando2013unsupervised,gong2012geodesic}, or distribution alignment to reduce the marginal or conditional distribution divergence between domains~\cite{wang2019easy,wang2018stratified,wang2017balanced,pan2011domain,}. Recently, deep transfer learning methods have made considerable success in many application fields~\cite{rozantsev2019beyond,ganin2014unsupervised,tzeng2014deep}. For a complete survey, please refer to \cite{pan2010survey}. 

FedHealth is mainly related to deep transfer learning. Most of the methods assume the availability of training data, which is not realistic. FedHealth makes it possible to do deep transfer learning in the federated learning framework without accessing the raw user data. Therefore, it is more secure.

\begin{figure*}[t!]
	\centering
	\includegraphics[scale=.5]{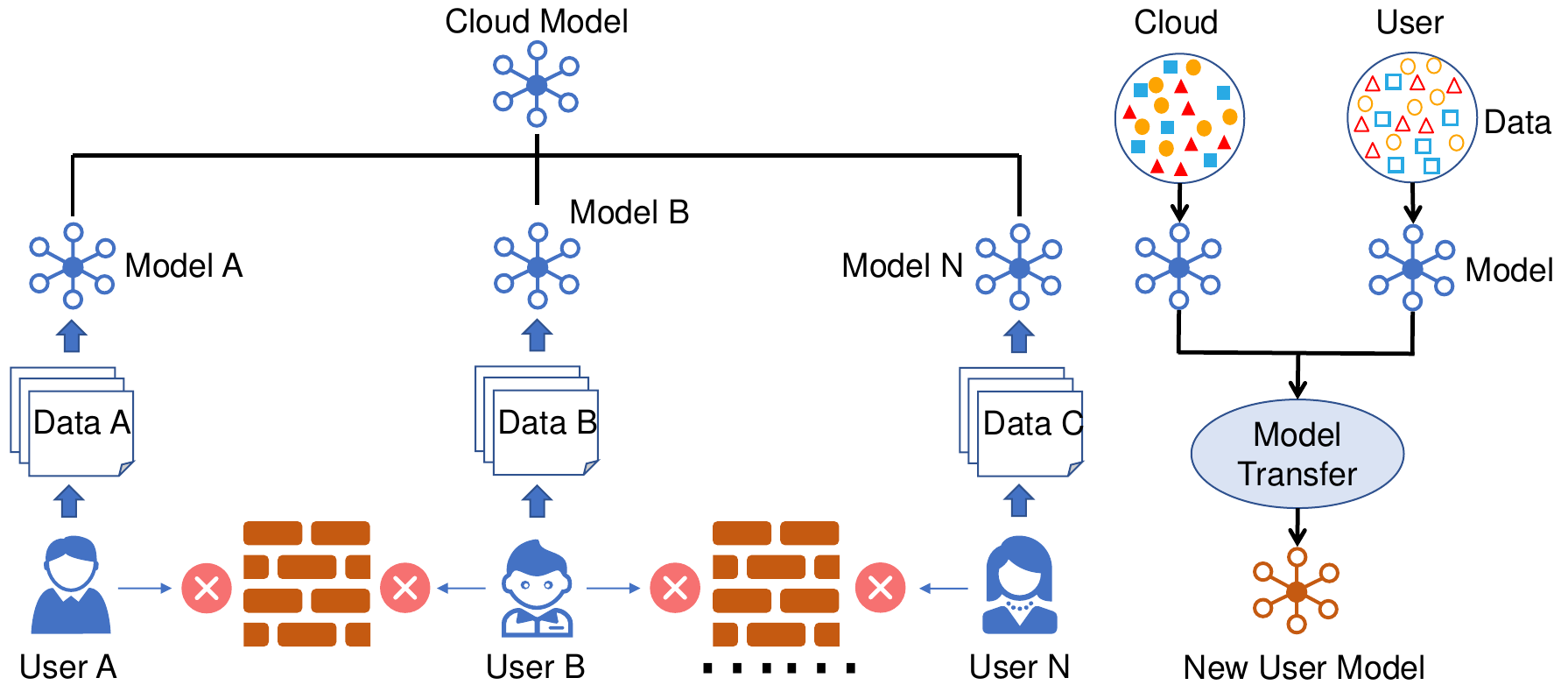}
	\caption{Overview of the FedHealth framework. ``User'' represents organizations}
	\label{fig-framework}
\end{figure*}

\section{The Proposed FedHealth Framework}

In this section, we introduce the FedHealth framework for federated transfer learning based wearable healthcare.

\subsection{Problem Definition}

We are given data from $N$ different users (organizations), denoted the users by $\{\mathcal{S}_1, \mathcal{S}_2,\cdots,\mathcal{S}_N\}$ and the sensor readings they provide are denoted by $\{\mathcal{D}_1, \mathcal{D}_2,\cdots,\mathcal{D}_N\}$. Conventional methods train a model $\mathcal{M}_{ALL}$ by combining all the data $\mathcal{D} = \mathcal{D}_1 \cup \mathcal{D}_2 \cup \cdots \cup \mathcal{D}_N$. All the data have different distributions. In our problem, we want to collaborate all the data to train a federated model $\mathcal{M}_{FED}$, where any user $\mathcal{S}_i$ does not expose its data $\mathcal{D}_i$ to each other. If we denote the accuracy as $\mathcal{A}$, then the objective of FedHealth is to ensure the accuracy of federated learning is close to that of conventional learning denoted by:
\begin{equation}
	|\mathcal{A}_{FED} - \mathcal{A}_{ALL}| < \Delta,
\end{equation}
where $\Delta$ is an extremely small non-negative real number.

\subsection{Overview of the Framework}

FedHealth aims to achieve accurate personal healthcare through federated transfer learning without compromising privacy security. Figure~\ref{fig-framework} gives an overview of the framework. Without loss of generality, we assume there are 3 users (organizations) and 1 server, which can be extended to the more general case. The framework mainly consists of four procedures. First of all, the cloud model on the server end is train based on public datasets. Then, the cloud model is distributed to all users where each of them can train their own model on their data. Subsequently, the user model can be uploaded to the cloud to help train a new cloud model. Note that this step does not share any user data or information but the encrypted model parameters. Finally, each user can train personalized models by integrating the cloud model and its previous model and data for personalization. In this step, since there is large distribution divergence between cloud and user model, transfer learning is performed to make the model more tailored to the user (right part in Figure~\ref{fig-framework}). It is noteworthy that all the parameter sharing processes does not involve any leakage of user data. Instead, they are finished through homomorphic encryption~\cite{rivest1978data}.

The federated learning paradigm is the main computing model for the whole FedHealth framework. It deals with model building and parameter sharing during the entire process. After the server model is learned, it can be directly applied to the user. This is just what traditional healthcare applications do for model learning. It is obvious that the samples in the server are having highly different probability distribution with the data generated by each user. Therefore, the common model fails in personalization. Additionally, user models cannot easily be updated continuously due to the privacy security issue.

\subsection{Federated Learning}

FedHealth adopts the federated learning paradigm~\cite{yang2019federated} to achieve encrypted model training and sharing. This step mainly consists of two critical parts: cloud and user model learning. After obtaining the server model, it is distributed to the user end to help them train their own models. As for each user, it trains its own model with the help of the server model.

In FedHealth, we adopt deep neural networks to learn both the cloud and user model. Deep networks perform end-to-end feature learning and classifier training by taking the raw inputs of the user data as inputs. Let $f_S$ denote the server model to be learned, then the learning objective becomes:
\begin{equation}
\label{eq-server}
\arg \min_{\Theta} \mathcal{L}= \sum_{i=1}^{n} \ell (y_i,f_S(\mathbf{x}_i)),
\end{equation}
where $\ell(\cdot,\cdot)$ denotes the loss for the network, e.g. cross-entropy loss for classification tasks. $\{\mathbf{x}_i,y_i\}^{n}_{i=1}$ are samples from the server data with $n$ their sizes. $\Theta$ denotes all the parameters to be learned, i.e. the weight and bias.

After acquiring the cloud model, it is distributed to all the users. As we can see from the ``wall'' in Figure~\ref{fig-framework}, direct sharing of user information is forbidden. This process uses homomorphic encryption~\cite{rivest1978data} to avoid information leakage. Since the encryption is not our main contribution, we will show the process of additively homomorphic encryption using real numbers. The encryption scheme of the weight matrix and bias vector are following the same idea. The additively homomorphic encryption of a real number $a$ is denoted as $\langle a \rangle$. In additively homomorphic encryption, for any two numbers $a$ and $b$, we have $\langle a \rangle + \langle b \rangle=\langle a+b \rangle$. Therefore, the parameter sharing can be done without leaking any information from the users. Through federated learning, we can aggregate user data without compromising privacy security.

Technically, the learning objective for user $u$ is denoted as:
\begin{equation}
	\label{eq-user}
	\arg \min_{\Theta^u} \mathcal{L}_1= \sum_{i=1}^{n^u} \ell (y^u_i,f_u(\mathbf{x}^u_i)).
\end{equation}

It is important to note that FedHealth does not perform parameter sharing as in \cite{Cheng2019SecureBoostAL} for computational efficiency. After all the user model $f_u$ is trained, it is uploaded to the server for aggregation. As for aggregation, server can align the old model with the model from each user subsequently. Considering the computational burden, server can also achieve scheduled update (e.g. every night) using uploading user models. The result is a new server model $f_S'$. Note that the new server model $f_S'$ is based on the knowledge from all users. Therefore, it has better generalization ability.

\subsection{Transfer Learning}

Federated learning solves the data islanding problem. Therefore, we can build models using all the user data. Another important factor is the personalization. Even if we can directly use the cloud model, it still performs poor on a particular user. This is due to the distribution difference between the user and the cloud data. The common model in the server only learns the coarse features from all users, while it fails in learning the fine-grained information on a particular user.

In this paper, FedHealth uses transfer learning to build a personalized model for each user. Recall that features in deep networks are highly transferable in the lower levels of the network since they focus on learning common and low-level features. The higher layers learn more specific features to the task~\cite{yosinski2014transferable}. In this way, after obtaining the parameters of the cloud model, we can perform transfer learning on the user to learn their personalized models.

\begin{figure}[t!]
	\centering
	\includegraphics[scale=.42]{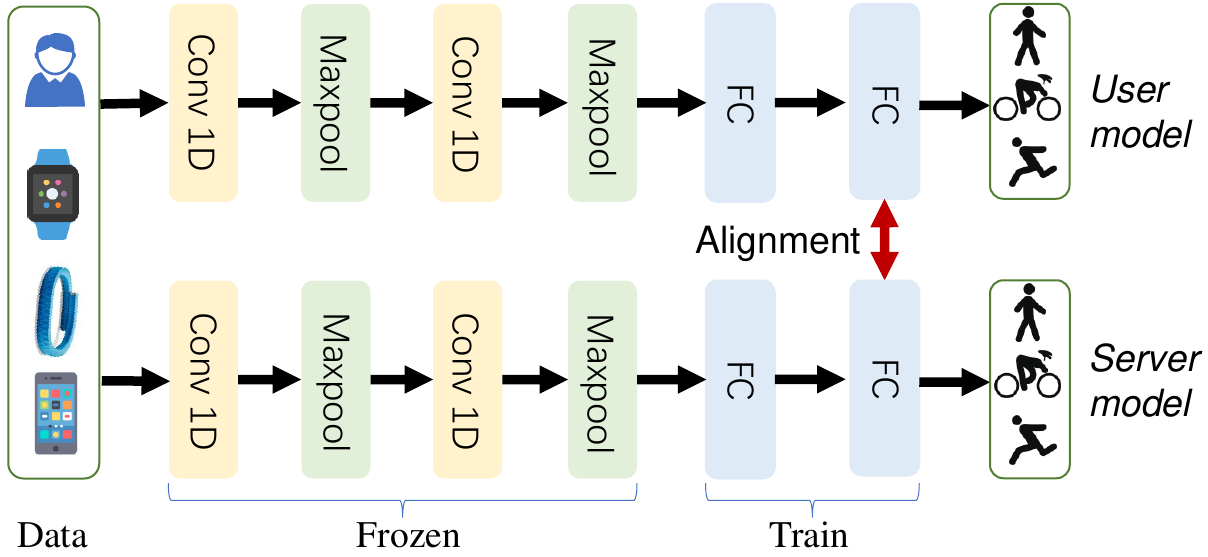}
	\caption{The transfer learning process of FedHealth}
	\vspace{-.2in}
	\label{fig-finetune}
\end{figure}

Figure~\ref{fig-finetune} presents the process of transfer learning for a specific convolutional neural network ~(CNN). Suppose the network is composed of two convolutions layers (\texttt{conv1, conv2}), two max-pooling layers (\texttt{pool1, pool2}), two fully connected layers (\texttt{fc1, fc2}), and one softmax layer for classification. The network is designed for human activity recognition where the input data is the activity signals for a user and the output is his activity classes. 

In model transfer, we think that the convolution layers aims at extracting low-level features about activity recognition. Thus we keep these layers along with the max-pooling layers frozen, which means we do not update their parameters in backpropagation. As for the fully connected layers \texttt{fc1} and \texttt{fc2}, since they are higher level, we believe they focus on learning specific features for the task and user. Therefore, we update their parameters during training. The softmax serves as the classification function, which can be formulated as:
\begin{equation}
	\label{eq-softmax}
	y_{j}=\frac{e^{z_{c}}}{\sum_{c=1}^{C} e^{z_{c}}},
\end{equation}
where $z_{c}$ denotes the learned probability for class $C$, and $y_{j}$ is the final classification result.

\begin{algorithm}[t!]
	\caption{The learning procedure of FedHealth}  
	\label{algo-fedhealth}  
	\renewcommand{\algorithmicrequire}{\textbf{Input:}} 
	\renewcommand{\algorithmicensure}{\textbf{Output:}}
	\begin{algorithmic}[1]  
		\REQUIRE 
		Data from different users $\{\mathcal{D}_1, \mathcal{D}_2,\cdots,\mathcal{D}_N\}$, $\eta$\\
		\ENSURE 
		Personalized user model $f_u$\\
		\STATE Construct a cloud model $f_S$ using Eq.~(\ref{eq-server})
		\STATE Distribute $f_S$ to all users via homomorphic encryption
		\STATE Train user models using Eq.~(\ref{eq-user})
		\STATE Update all user models to the server using homomorphic encryption. Then server update its model by aligning with user model
		\STATE Distribute $f_S'$ to all users, then perform transfer learning on each user to get their personalized model $f_u$ using Eq.~(\ref{eq-usercoral})
		\STATE Repeat the above procedures with the continuously emerging user data
	\end{algorithmic}
\end{algorithm}

FedHealth adapts the inputs from different domains by replacing \texttt{fc2} with an \texttt{alignment} layer. This is strictly different that in DDC~\cite{tzeng2014deep} and other recent methods where we have access to both the source and target data. In our problem, we only have the user data and the cloud model. To this end, we borrow the idea from \cite{rozantsev2018BeyondSW} and regularize the weights. Given the network from the server and user, we add a correlation alignment~\cite{sun2016return} layer before the softmax layer to further adapt the domains. This alignment function is used to align the second-order statistics between the inputs. Formally, the loss of correlation alignment is computed as follows:
\begin{equation}
\label{eq-coral}
\ell_{C O R A L}=\frac{1}{4 d^{2}}\left\|C_{S}-C_{T}\right\|_{F}^{2}
\end{equation}
where $\left\| \cdot \right\|_F^{2}$ denotes the squared matrix Frobenius norm and $d$ is the dimension of the embedding features. $C_S$ and $C_T$ are the covariance matrices of the source and target weights computed by \cite{sun2016return}. Therefore, denote $\eta$ the trade-off parameter, the loss for the user model is computed by:
\begin{equation}
\label{eq-usercoral}
	\arg \min_{\Theta_u} \mathcal{L}_u= \sum_{i=1}^{n_u} \ell (y^u_i,f_u(\mathbf{x}^u_i)) + \eta \ell_{CORAL}.
\end{equation}

\subsection{Learning Process}

The learning procedure of FedHealth is presented in Algorithm~\ref{algo-fedhealth}. Note that this framework works continuously with the new emerging user data. FedHealth can update the user model and cloud model simultaneously when facing new user data. Therefore, the longer the user uses the product, the more personalized the model can be. Other than transfer learning, FedHealth can also embed other popular methods for personalization such as incremental learning~\cite{rebuffi_2017_CVPR}. 

The entire framework can also adopt other machine learning methods other than deep networks. For instance, the gradient boosting decision tree can be integrated into the framework to harness the power of ensemble learning. These lightweight models can be deployed to computation restricted wearable devices. This makes FedHealth more general to real applications.

\section{Experiments}

In this section, we evaluate the performance of the proposed FedHealth framework via extensive experiments on human activity recognition.

\subsection{Datasets}

We adopt a public human activity recognition dataset called UCI Smartphone~\cite{anguita2012human}. This dataset contains 6 activities collected from 30 users. The 6 activities are WALKING, WALKING-UPSTAIRS, WALKING-DOWNSTAIRS, SITTING, STANDING, and LAYING. There are 30 volunteers within an age bracket of 19-48 years. Each volunteer wears a smartphone (Samsung Galaxy S II) on the waist. Using its embedded accelerometer and gyroscope, collectors captured 3-axial linear acceleration and 3-axial angular velocity at a constant rate of 50Hz. The experiments have been video-recorded to label the data manually. The obtained dataset has been randomly partitioned into two sets, where 70\% of the volunteers were selected for generating the training data and 30\% the test data. There are 10,299 instances in total. The statistical information of the dataset is shown in Table~\ref{tb-dataset}.

\begin{table}[htbp]
	\caption{Statistical information of the dataset}
	\label{tb-dataset}
	\resizebox{.48\textwidth}{!}{
	{}
	\begin{tabular}{cccccc}
		\toprule
		Subject & Activity & Sampling rate & Sensor & Instance & Channel \\ \hline
		30 & 6 & 50 Hz & \begin{tabular}[c]{@{}c@{}}Accelerometer\\ Gyroscope\end{tabular} & 10,299 & 9 \\ \bottomrule
	\end{tabular}
}
\end{table}

In order to construct the problem situation in FedHealth, we change the standard setting for the dataset. We extracted 5 subjects (Subject IDS 26 $\sim$ 30) and regard them as the isolated users which cannot share data due to privacy security. Data on the remaining 25 users are used to train the cloud model. Henceforth, the objective is to use the cloud model and all the 5 isolated subjects to improve the activity recognition accuracy on the 5 subjects without compromising the privacy. In short, it is a variant of the framework in Figure~\ref{fig-framework} where there are 5 users.

\subsection{Implementation Details}

On both the server and the user end, we adopt a CNN for training and prediction. The network is composed of 2 convolutional layers, 2 pooling layers, and 3 fully connected layers. The network adopts a convolution size of $1 \times 9$. It uses mini-batch Stochastic Gradient Descent (SGD) for optimization. During training, we use 70\% of the training data for model training, while the rest 30\% is for model evaluation. We fix $\eta=0.01$. We set the learning rate to be 0.01 with batch size of 64 and training epochs fixed to 80. The accuracy of user $u$ is computed as $\mathcal{A}_u = \frac{\left|\mathbf{x} : \mathbf{x} \in \mathcal{D}_{u} \wedge \hat{y}(\mathbf{x})=y(\mathbf{x})\right|}{\left|\mathbf{x} : \mathbf{x} \in \mathcal{D}_{u}\right|}$, where $y(\mathbf{x})$ and $\hat{y}(\mathbf{x})$ denote the true and predicted labels on sample $\mathbf{x}$, respectively.

We follow~\cite{rivest1978data} for homomorphic encryption in federated learning. During transfer learning, we freeze all the convolutional and pooling layers in the network. Only the parameters of the fully connected layers are updated with SGD. To show the effectiveness of FedHealth, we compare its performance with traditional learning, where we record the performances on each subject using the server model only. For notational brevity, we use \texttt{NoFed} to denote this setting. We also compare the performances of KNN, SVM, and random forest (RF) with FedHealth. The hyperparameters of all the comparison methods are tuned using cross-validation. For the fair study, we run all the experiments 5 times to record the average accuracies. 

\subsection{Classification Accuracy}

The classification accuracies of activity recognition for each subject are shown in Table~\ref{tb-acc}. The results indicate that our proposed FedHealth achieves the best classification accuracy on all users. Compared to NoFed, it significantly improves the average results by \textbf{5.3\%}. Compared to the traditional methods~(KNN, SVM, and RF), FedHealth also greatly improves the recognition results. In short, it demonstrates the effectiveness of our proposed FedHealth framework.

\begin{table}[t!]
	\caption{Classification accuracy~(\%) of the test subject}
	\label{tb-acc}
	\resizebox{.48\textwidth}{!}{
	\begin{tabular}{cccccc}
		\toprule
		Subject & KNN & SVM & RF & NoFed & FedHealth \\ \hline
		P1 & 83.8 & 81.9 & 87.5 & 94.5 & \textbf{98.8} \\ 
		P2 & 86.5 & 96.9 & 93.3 & 94.5 & \textbf{98.8} \\ 
		P3 & 92.2 & 97.2 & 88.9 & 93.4 & \textbf{100.0} \\ 
		P4 & 83.1 & 95.9 & 91.0 & 95.5 & \textbf{99.4} \\ 
		P5 & 90.5 & 98.6 & 91.6 & 92.6 & \textbf{100.0 }\\ 
		AVG & 87.2 & 94.1 & 90.5 & 94.1 & \textbf{99.4} \\ \bottomrule
	\end{tabular}
}
\end{table}

\begin{figure}[t!]
    \hspace{-.1in}
	\includegraphics[scale=.36]{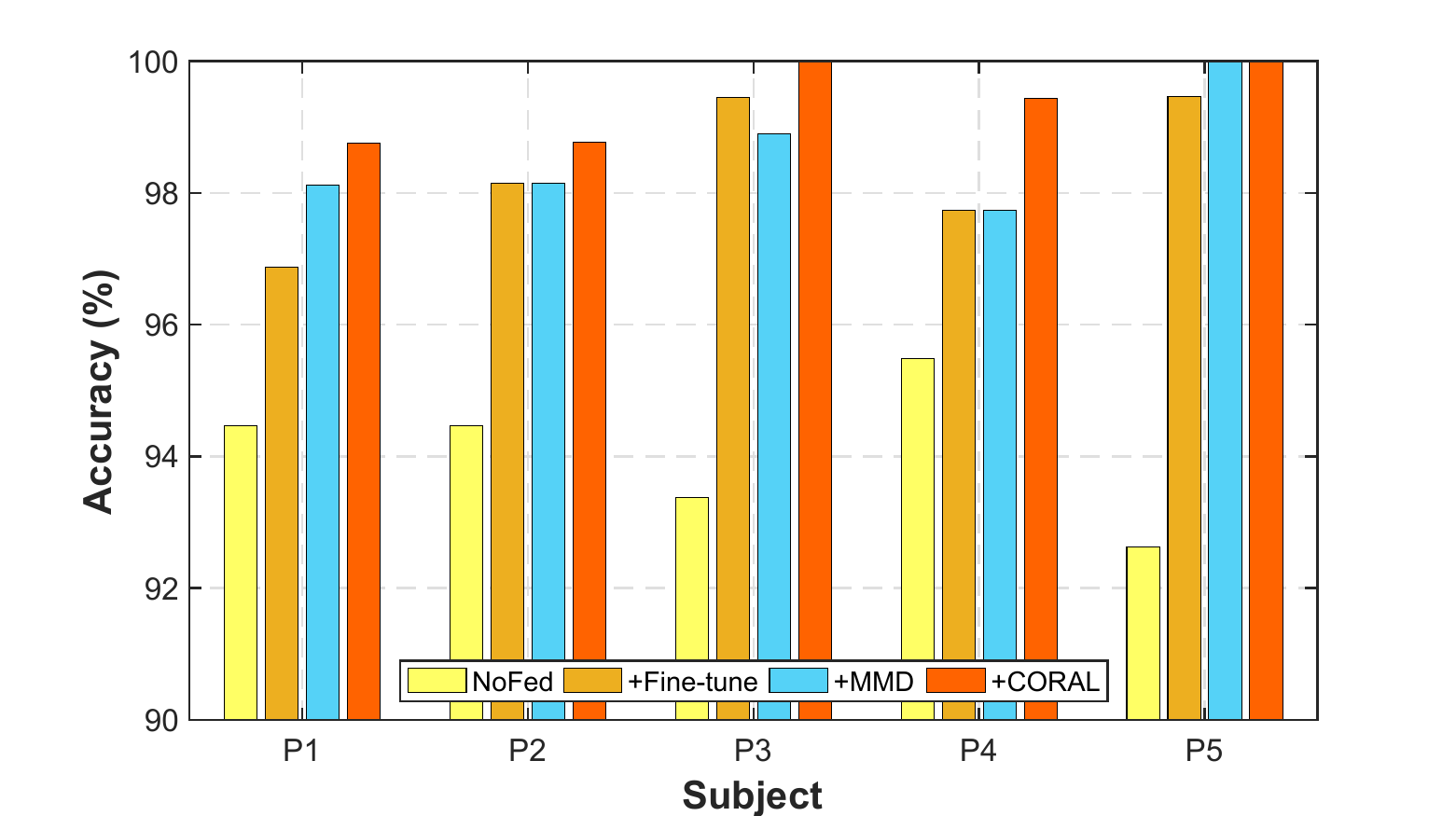}
	\caption{Extending FedHealth with other transfer learning methods}
	\label{fig-ablation}
\end{figure}

The results also show that for activity recognition, the deep methods (NoFed and FedHealth) achieve better results than traditional methods. This is due to the representation capability of deep neural networks, while traditional methods have to rely on hand-crafted feature learning. Another advantage of deep learning is that the models can be updated online and incrementally without retraining, while traditional methods require further incremental algorithms. This property is extremely valuable in federated transfer learning where model reuse is important and helpful.

\subsection{Evaluation of Extensibility}

In this section, we analyze the extensibility of FedHealth with different transfer learning approaches. We compare its performance with two methods: 1) fine-tuning, which only fine-tunes the network on each subject without explicitly reducing the distribution divergence between domains; and 2) transfer with MMD (Maximum Mean Discrepancy)~\cite{wang2018visual}, which replaces the alignment loss with MMD loss. The comparison results are shown in Figure~\ref{fig-ablation}. 

From the results, we can see that other than the alignment loss, FedHealth can also achieve promising results using fine-tuning or MMD. The results of transfer learning significantly outperform no transfer by 4\% on average accuracy. This indicates that the transfer learning procedure of FedHealth is highly effective and extensible. Therefore, FedHealth is general and can be extended in many applications by integrating other transfer learning algorithms. Moreover, the federated learning procedure can also be extended using other encryption methods, which can be the future research.

\subsection{Detailed Analysis}

We provide detailed analysis to FedHealth via comparing its confusion matrix with that of NoFed. The confusion matrix is known as an effective metric to show the efficacy of a method since it provides fine-grained classification results on each task. For simplicity, we show the confusion matrices of subject 2 in Table~\ref{tb-conf}. The results of other subjects follow the same tendency. Along with the confusion matrix, the precision~($P$), recall~($R$), and macro $F1$ score~($F1$) are all computed to give a thorough view of the results.

Combining the results in Table~\ref{tb-acc} and \ref{tb-conf}, we can clearly see that FedHealth can not only achieve the best accuracy, but also reach the best precision, recall, and $F1$ scores. The confusion matrix shows that FedHealth can reduce the misclassification rate, especially on class $C1$ (Walking). Since walking is the most common activities in healthcare, it means that FedHealth is effective in recognizing this activity. To summarize, FedHealth is more accurate in recognizing personalized activities, which makes it more advantageous in healthcare applications.

\section{Discussions}

FedHealth is a general framework for wearable healthcare. This paper provides a specific implementation and evaluation of this idea. It is adaptable to several healthcare applications. In this section, we discuss its potential to be extended and deployed to other situations with possible solutions.

1. FedHealth with incremental learning. Incremental learning~\cite{rebuffi_2017_CVPR} has the ability to update the model with the gradually changing time, environment, and users. In contrast to transfer learning that focuses on model adaptation, incremental learning makes it possible to update the model in real-time without much computation.

2. FedHealth as the standard for wearable healthcare in the future. FedHealth provides such a platform where all the companies can safely share data and train models. In the future, we expect that FedHealth be implemeted with blockchain technology~\cite{zheng2018blockchain} where user data can be more securely stored and protected. We hope that FedHealth can become the standard for wearable healthcare.

3. FedHealth to be applied in more applications. This work mainly focuses on the possibility of federated transfer learning in healthcare via activity recognition. In real situations, FedHealth can be deployed at large-scale to more healthcare applications such as elderly care, fall detection, cognitive disease detection, etc. We hope that through FedHealth, federated learning can become federated computing which can become a new computing model in the future.

\begin{table}[t!]
	\caption{Classification report of NoFed and FedHealth}
	\label{tb-conf}
	\resizebox{.48\textwidth}{!}{
		\begin{tabular}{ccccccc|ccc}
			\hline
			\multicolumn{10}{c}{\textbf{NoFed}} \\ 
			& $C1$ & $C2$ & $C3$ & $C4$ & $C5$ & $C6$ & $P$ & $R$ & $F1$ \\ \hline
			$C1$ & 70.4\% & 7.4\% & 22.2\% &  &  &  & 1 & 0.7 & 0.83 \\ 
			$C2$ &  & 96.9\% & 3.1\% &  &  &  & 0.94 & 0.97 & 0.95 \\ 
			$C3$ &  &  & 100\% &  &  &  & 0.74 & 1 & 0.85 \\ 
			$C4$ &  &  &  & 100\% &  &  & 1 & 1 & 1 \\
			$C5$ &  &  &  &  & 100\% &  & 1 & 1 & 1 \\ 
			$C6$ &  &  &  &  &  & 100\% & 1 & 1 & 1 \\ 
			\multicolumn{7}{c|}{Average} & 0.96 & 0.94 & 0.94 \\ \hline \hline
			\multicolumn{10}{c}{\textbf{FedHealth}} \\ 
			& $C1$ & $C2$ & $C3$ & $C4$ & $C5$ & $C6$ & $P$ & $R$ & $F1$ \\ \hline
			$C1$ & \textbf{88.9\%} & 3.7\% & 7.4\% &  &  &  & 1 & \textbf{0.89} & \textbf{0.94} \\ 
			$C2$ &  & \textbf{100\%} &  &  &  &  & \textbf{0.97} & 1 & \textbf{0.98} \\ 
			$C3$ &  &  & 100\% &  &  &  & \textbf{0.91} & 1 & \textbf{0.95} \\
			$C4$ &  &  &  & 100\% &  &  & 1 & 1 & 1 \\ 
			$C5$ &  &  &  &  & 100\% &  & 1 & 1 & 1 \\ 
			$C6$ &  &  &  &  &  & 100\% & 1 & 1 & 1 \\ 
			\multicolumn{7}{c|}{Average} & \textbf{0.98} & \textbf{0.98} & \textbf{0.98} \\ \hline
		\end{tabular}
	}
\end{table}

\section{Conclusions and Future Work}

In this paper, we propose FedHealth, the first federated transfer learning framework for wearable healthcare. FedHealth aggregates the data from different organizations without compromising privacy security, and achieves personalized model learning through knowledge transfer. Experiments on human activity recognition have demonstrated the effectiveness of the framework. We also present a detailed discussion for its potential from specific technical improvement to the potential for healthcare applications.

FedHealth opens a new door for future research in wearable healthcare. In the future, we plan to extend FedHealth to the detection of Parkinson's disease where it can be deployed in hospitals.

\section*{Acknowledgements}
This paper is supported in part by National Key R \& D Plan of China (No. 2017YFB1002802), NSFC (No. 61572471), and Beijing Municipal Science \& Technology Commission~(No.Z171100000117017).

\bibliographystyle{named}
\bibliography{ijcai19}

\end{document}